\pdfoutput=1
\documentclass[11pt]{article}
\usepackage{adjustbox}
\usepackage[inline]{enumitem}
\usepackage{EMNLP2023}
\usepackage{times}
\usepackage{latexsym}
\usepackage{comment}
\usepackage{graphicx}
\usepackage{graphics}
\usepackage{epigraph}
\usepackage[utf8]{inputenc}

\usepackage{verbatim}
\usepackage{graphicx}

\mathchardef\mhyphen="2D
\usepackage{xcolor}
\usepackage{enumitem}

\usepackage{svg}
\usepackage{float}
\usepackage{hyperref}
\usepackage{amsmath}
\usepackage{amssymb} 
\usepackage{float}
\usepackage{xspace}
\usepackage[T1]{fontenc}
\usepackage[utf8]{inputenc}

\usepackage{microtype}

\usepackage{inconsolata}
\usepackage[color=green]{todonotes}
\usepackage{url}
\usepackage{booktabs}
\usepackage{enumitem}
\usepackage{multirow}
\usepackage{multicol}
\usepackage[ruled, vlined, noresetcount, linesnumbered]{algorithm2e}

\title{A Diachronic Analysis of Paradigm Shifts in NLP Research:\\ When, How, and Why?}

\author{
Aniket Pramanick\textsuperscript{1}, Yufang Hou\textsuperscript{2}, Saif M. Mohammad\textsuperscript{3}, Iryna Gurevych\textsuperscript{1} \\
\textsuperscript{1}Ubiquitous Knowledge Processing Lab (UKP Lab) \\
Department of Computer Science and Hessian Center for AI (hessian.AI) \protect\\
\textsuperscript{2}IBM Research Europe, Ireland \\
\textsuperscript{3}National Research Council Canada \\
\texttt{\url{www.ukp.tu-darmstadt.de}, \url{yhou@ie.ibm.com}, \url{saif.mohammad@nrc-cnrc.gc.ca}}
}

\begin{document}

\maketitle

\begin{abstract}

Understanding the fundamental concepts and trends in a scientific field is crucial for keeping abreast of its continuous advancement. In this study, we propose a systematic framework for analyzing the evolution of research topics in a scientific field using causal discovery and inference techniques. We define three variables to encompass diverse facets of the evolution of research topics within NLP and utilize a causal discovery algorithm to unveil the causal connections among these variables using observational data. Subsequently, we leverage this structure to measure the intensity of these relationships. By conducting extensive experiments on the \textit{ACL Anthology} corpus, we demonstrate that our framework effectively uncovers evolutionary trends and the underlying causes for a wide range of NLP research topics. Specifically, we show that tasks and methods are primary drivers of research in NLP, with datasets following, while metrics have minimal impact.
\footnote{We publish the code and dataset for our experiments at \url{https://github.com/UKPLab/CausalNLPTrends}}

\end{abstract}

\section{Introduction}
\label{sec:intro}

\begin{figure*}
    \centering
    \includegraphics[width=1\textwidth]{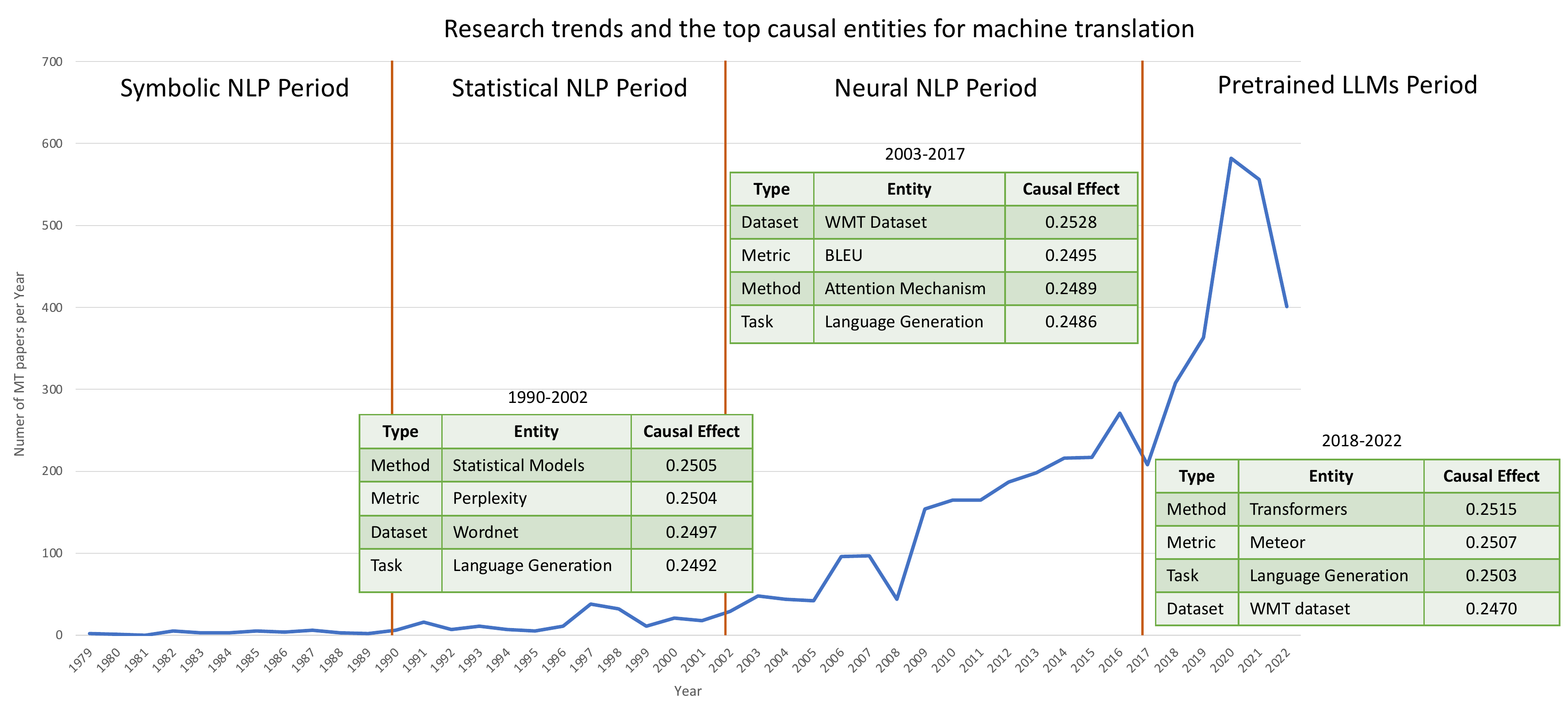}
    \caption{Evolution of Machine Translation (MT) research. Blue line: Number of MT papers (1979-2022). Tables show the top causal entities/types for different periods (excluding 1979-1989 due to limited MT papers).}
    %\caption{Evolution timeline of Machine Translation (MT). The blue line shows the number of papers on MT from 1979 to 2022. Tables summarize the top causal entities and their types for MT in different time periods. The causal analysis result on 1979-1989 is omitted due to a relatively small number of MT papers in this time period, in which we cannot obtain the statistically significant causal entities for MT.}
    \label{fig:nmt_development}
    %\vspace*{-3mm}
\end{figure*}

Experts in a field sometimes conduct historical studies to synthesize and document the key research ideas, topics of interest, methods, and datasets that shaped a field of study. 
They document how new research topics eclipsed older ones and contributed to shaping the trajectory of the research area \cite{sep-thomas-kuhn}. 
Aspiring scientists learn the craft of their discipline by delving into the examination of past scientific accomplishments documented in research papers. However, conducting such a historical study is challenging: Experts in a field rely on years of experience and peruse large amounts of past published articles 
% needs specialized knowledge as well as extensive analysis 
to determine
% of 
the chronological progression of a research field. Further, the exponential growth of scientific publications in recent years has rendered it arduous even for domain experts to stay current. Therefore, an automated method to track the temporal evolution of research topics can be beneficial in offering an overview of the field and assisting researchers in staying abreast of advancements more efficiently.
%becomes helpful in providing 

In this work, we propose a systematic framework to examine the evolutionary journey of research topics within the realm of Natural Language Processing (NLP), harnessing causal discovery and inference techniques. %While
Prior research on historical analysis of NLP has predominantly concentrated on scrutinizing metadata associated with research papers \cite{hall-etal-2008-studying,Mohammad2019TheSO, uban-etal-2021-studying,singh-etal-2023-forgotten,wahle-etal-2023-citation-field} such as number of citations, title, author profile, affiliation, and publication venue. These studies have examined the research trends through unigram or bigram frequency analysis, but they do not provide insights into the underlying causes propelling these research topics.

Our study centers on four distinct fundamental types of entities in NLP research: 
\emph{\bf tasks} representing well defined problems; \emph{\bf methods},  signifying the solutions or approaches employed to tackle the tasks; \emph{\bf datasets}, indicating the relevant textual resources such as corpora and lexicons; and \emph{\bf metrics}, encompassing the evaluation techniques tailored to specific tasks. %(\textbf{TDMM} for short). 
We abbreviate these types as \textbf{TDMM} for short. Specifically, we examine the interplay between an NLP task that is commonly viewed as a focused research topic (e.g., \emph{Machine Translation}) and the key \emph{entities} that exert pivotal influence on the target task (such as ``\emph{BLEU}'' \cite{papineni2002bleu} or ``\emph{Transformers}'' \cite{transformer}).

Our goal is to identify the TDMM entities ($E$) associated with a specific task ($t$) and assess their causal influence on the task's research trends (\textbf{TDMM-Task causal analysis}). Specifically, we address the following key research questions %when presented 
associated with a task entity $t$: (a) Which entities $E$ effectively indicate the research trends for this task $t$? (b) Are there discernible causal relationships between $t$ and $E$? (c) What is the extent of the causal impact exerted by $E$ on $t$?

Unlike \citet{uban-etal-2021-studying} and \citet{koch2021reduced} that heavily rely on manual annotations and have limited coverage, our analysis is based on TDMM 
%(\emph{Task/Dataset/Metric/Method}) 
entities automatically extracted from 55K  papers in the ACL Anthology\footnote{\url{https://aclanthology.org/}}. Our framework not only recognizes the key entities driving the research direction of a research topic but also measures the causal effects of these entities on %contemporary entities %on these topics 
the target topic
in an end-to-end fashion.
Figure \ref{fig:nmt_development} shows the most influential entities for \emph{Machine Translation} (MT) in different time periods. For instance, ``\emph{statistical models}'' used to be the popular method for MT in  1990-2002, and the evaluation metric ``\emph{BLEU}'' is one of the top causal entities driving the MT research in 2003-2017. In the era of pre-trained large language models (LLMs) starting from 2018, ``\emph{transformer}'' has become the popular method for MT.  
For another research topic of ``\emph{Speech recognition}'', our framework uncovers the influential role of ``\emph{language modeling}'' between 1979 to 2022, where speech recognition models utilize probability scores from language models to recognize coherent text from speech \citep{negri-etal-2014-quality}.
%Our framework also reveals the influential role of {\it ``language modeling''} in {\it ``Speech recognition''} research, where speech recognition models often utilize probability scores from language models to recognize coherent text from speech \citep{negri-etal-2014-quality}. 

In this work, we analyze 16 tasks from a diverse set of research areas identified by ACL 2018 organizers. Our framework is versatile and applicable to other tasks and domains, benefiting both young and experienced researchers. It can aid in literature surveys by identifying related research areas and enable %experienced 
young researchers to delve into new research focuses by establishing connections %to new knowledge. 
among different research areas.

%In this work, we study 16 tasks, each representing a different area categorized by the ACL 2018 organizers. Our framework can easily be applied to investigate other tasks and explore research in various domains. By leveraging our framework, both young and experienced researchers can benefit. It assists in literature surveys, aiding in the identification of related research areas while also enabling experienced researchers to delve into new research focuses by identifying key related areas and establishing connections to new knowledge.

In summary, we make \textit{three-fold} contributions in this study: {\bf Firstly}, we propose a framework to quantify research activities, including (1) trends and stability of an NLP research task, and (2) relation intensity between TDMM entities and NLP research tasks. {\bf Secondly}, we employ causal analysis algorithms to uncover causal structures and measure effects between tasks and related TDMM entities (\emph{TDMM-Task causal analysis}). To the best of our knowledge, this represents the first historical study of a scientific research anthology from a causal perspective. {\bf Finally}, through extensive experiments on the ACL Anthology, we offer an empirical overview of the NLP research landscape. In the following sections, we will refer to {\it TDMM-Task causal analysis} as {\it causal analysis}.

\begin{figure*}
    \centering
    \begin{adjustbox}{width=1.8\columnwidth, center}
    \includegraphics[width=1.0\textwidth]{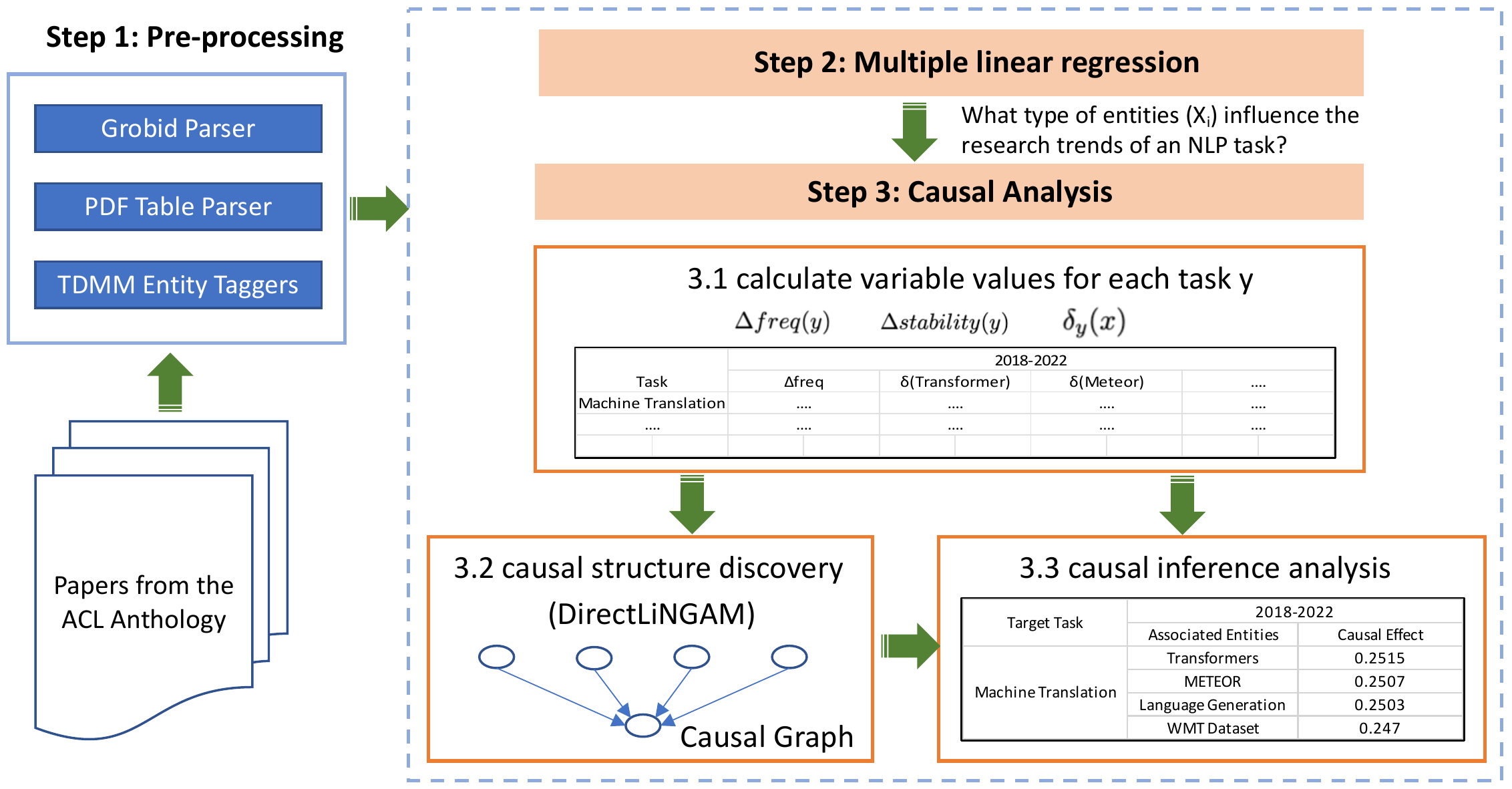}
    \end{adjustbox}
    \caption{System architecture.}
    \label{fig:method}
    %\vspace*{-3mm}
\end{figure*}

\section{Related Work}
\label{sec:rw}

%In addition to what we discussed in the previous section; we want to add two strands of related studies.
\paragraph{Scientific Trends Analysis}
%\subsection{Scientific Trends Analysis}
The analysis of scientific trends has been a research focus since \citet{hall-etal-2008-studying}. In the field of ``scientometrics'', extensive literature explores citation patterns and utilizes topological measures in citation networks for trend analysis \citep{small_et_al, SHIBATA2008758, 10.1162/qss_a_00202}.

Another line of research focuses on metadata and content analysis. For instance, \citet{prabhakaran-etal-2016-predicting} employed rhetorical framing to examine trend patterns. \citet{Grudin_2009}, \citet{Liu2015ExploringTL}, and \citet{Mohammad2019TheSO} investigated the interaction between the topics in publications, research grants, author profiles, highly impactful papers, and dataset usage patterns. Additionally, \citet{koch2021reduced} studied dataset usage patterns among different research communities, while \citet{uban-etal-2021-studying} analyzed relationships between NLP research topics based on their co-occurrence in text and the degree of correlation between their popularity over time. In our work, we develop entity recognition models to extract TDMM entities from NLP research papers and focus on analyzing the causal relations between a task entity and its related TDMM entities.

\paragraph{Causality in NLP}
%\subsection{Causality in NLP}
%\citet{feder-etal-2022-causal} provides a survey about  an 
Existing works on NLP applying causal analysis algorithms mainly focus on two directions. The first line of work discovers causal relations among textual features or expressions of events in texts and uses them in various downstream tasks, such as question answering \citep{Oh_Torisawa_Hashimoto_Iida_Tanaka_Kloetzer_2016}, commonsense reasoning \citep{bosselut-etal-2019-comet, Sap_Le_Bras_Allaway_Bhagavatula_Lourie_Rashkin_Roof_Smith_Choi_2019}, and relation extraction \citep{do-etal-2011-minimally, mirza-tonelli-2014-analysis, dunietz-etal-2017-corpus}. 
%Please refer to the survey paper by ~\citet{feder-etal-2022-causal} for more details.

In another avenue of this field, researchers represent causal elements using textual features \citep{jin-etal-2021-mining-cause, fong-grimmer-2016-discovery, pmlr-v124-veitch20a, keith-etal-2020-text} and define the causal graph structure based on domain knowledge.
%In another avenue of this field, researchers use textual features to represent elements of the causal graph, e.g., the cause \citep{jin-etal-2021-mining-cause}, the effect \cite{fong-grimmer-2016-discovery} and the confounders \citep{pmlr-v124-veitch20a, keith-etal-2020-text}; they then use domain knowledge to define the structure of the causal graph for inference. 
Our work falls within this line of research, where we employ causal algorithms to analyze the trends in NLP research topics and the underlying causes.

\section{Data Collection}
\label{sec:dataset}

\begin{table*}[t!]
    \begin{center}
    %\resizebox{2.09\columnwidth}{!}{
    \begin{adjustbox}{width=2.0\columnwidth, center}
        \begin{tabular}{p{4.0cm} c p{12.0cm}}
        \toprule
        Period & Years & Key Research Themes \\
        \midrule
         Early Years & 1979--1989 & Foundational work in syntactic parsing, machine translation, and information retrieval. \\

         \midrule
         
         Formative Years & 1990--2002 & Advances in language modeling, named entity recognition, and discourse analysis (research focus shifted towards data-driven approaches). \\

         \midrule
         
         Statistical Revolution \& Neural Networks & 2003--2017 & Focus on statistical techniques (text classification, statistical machine translation, etc.) and resurgence of neural networks (word embeddings, neural machine translation, etc.) \\

         \midrule
        
         Deep Learning Era & 2018--2022 & Dominance of transformer-based architectures (BERT and their variants). \\
         
         \bottomrule
         % \bottomrule
         
        \end{tabular}
        %}
    \end{adjustbox}  
        \caption{Chronological Periods of NLP Research.}
        \label{tab:year_intervals}  
    \end{center}
    \vspace*{-3mm}
\end{table*}

\paragraph{ACL Anthology Corpus}
Following prior work by \citet{mohammad-2020-examining}, we utilize ACL Anthology as the source of NLP Research papers.
%ACL Anthology is a rich source of NLP research papers. 
For this work, we collect 55,366 NLP papers that belong to the ``ACL Events'' category\footnote{This category covers major NLP conferences, workshops, and journals including ACL, NAACL, EMNLP, EACL, AACL, CL, and TACL. Additionally, we also include papers published at COLING from the ``Non-ACL events category''.} from the ACL anthology published between 1979 and 2022. For each paper, we use GROBID \cite{GROBID} and the PDF table parser from \citet{hou-etal-2019-identification} to extract sentences from each of the individual sections as well as from the table and figure captions. In a post-processing step, we remove all the URLs from the extracted sentences. On average, we have 1,258 papers per year and 1,117 sentences per paper. %Please refer to Appendix \ref{sec:appendixa} for year-wise data statistics.

It is worth noting that certain NLP paper preprints may become accessible on preprint servers before they are officially published in the ACL Anthology. However, we argue that the peer review process in ACL Anthology serves as a robust quality assurance mechanism. Hence, we consider ACL Anthology a more reliable source compared to preprint servers. 

\paragraph{TDMM Entity Extraction}
To identify %mentions of 
{\it tasks}, {\it datasets}, {\it metrics}, and {\it methods} entities from %these sentences,
NLP papers, we developed two entity taggers based on Flair \citep{akbik-etal-2018-contextual}. The first tagger is based on the TDMSci annotations \citep{hou-etal-2021-tdmsci} for recognizing {\it task}, {\it dataset}, and {\it metric} entities. The second tagger is trained using the SciERC dataset \citep{luan-etal-2018-multi} to extract {\it method} entities. On the testing datasets of TDMSci and SciERC, the two taggers achieve a micro-average F1 of 0.77 and 0.78 for the type partial match \citep{segura-bedmar-etal-2013-semeval}, respectively. In type partial match, a predicted entity is considered correct if it partially overlaps with a gold entity and has the same type. For example, ``\emph{Penn Treebank}'' is counted as a correct prediction even if the corresponding gold annotation is ``\emph{Penn Treebank dataset}''.

To further improve the precision of the TDMM taggers, we include only entities that appear in more than five papers in the dataset. For each paper, we collect the most frequent task mentions appearing in the {\it title}, {\it abstract}, {\it experiment section}, {\it table}, and {\it figure captions} to approximate the tasks that the paper has done research on. 
 
\paragraph{Taxonomy for Periods of Reference} In order to facilitate in-depth analysis, in this paper, we adopt a taxonomy that partitions our reference time frame (1979-2022) into four distinct intervals. Table~\ref{tab:year_intervals} illustrates the defined intervals. These intervals have been designed to approximate the overarching trends observed in NLP research throughout the years, aligning with our perspective on the field's evolution. It is important to acknowledge that the exact boundaries and thematic emphases may differ based on varying perspectives and specific research areas within NLP. However, we highlight that our framework and methodologies are highly adaptable, allowing end users to effortlessly apply them to any desired time interval or a specific analysis.

\section{Entity Influence in NLP Research: A Regression Analysis}
\label{subsec:mult_reg}

\begin{table}[t]
    \begin{center}
    \begin{adjustbox}{width=0.7\columnwidth,center}
    \begin{tabular}{l c}
        \toprule
        
        Variables &  R-Squared ($\uparrow$)\\
        
        \midrule
        
         unique tasks & 0.87 \\
         + unique datasets & 0.91 \\
         + unique methods & 0.93 \\
         + unique metrics & {\bf 0.97} \\
                 \bottomrule
    \end{tabular}
    \end{adjustbox}
    \caption{{ Variable Selection for Regression.} 
    %Results show that the model fits well when we use all four variables - number of unique tasks, datasets, methods and metrics are used together. 
    %For details refer Section~\ref{subsec:regression}.
    }
    \label{tab:reg_var}
    \end{center}
    \vspace*{-3mm}
\end{table}

Before conducting the causal analysis, we aim to identify the key variables that significantly impact the evolution of NLP Research. Specifically, we investigate which types of entities exert the most influence on the research direction of NLP. To achieve this understanding, we employ Multiple Linear Regression (see Appendix~\ref{sec:mult_lin_reg} for details), a widely utilized tool in economics research \citep{NBERw27008}. Figure~\ref{fig:method} (step1/step2) illustrates the framework.

Our analysis assumes that if the TDMM entities have played a role in the emergence or disappearance of task entities, this influence will be reflected in the number of unique task entities in subsequent years, which can be captured through regression analysis. While the study does not provide specific information on the precise influence of each TDMM entity on individual task entities, the partial regression coefficients shed light on the types of entities responsible for influencing the overall task entity landscape.

%\subsection{Method}
\paragraph{Method.} 
Mathematically, we predict the number of task entities $Y^t$ in a given year $t$ as a function of the cumulative counts of all types of entities $\{X_{i}^{t_{-1}}\}$ (TDMM entities) until that year, $t_{-1}$, given by $Y^t = r_0 + \sum_{i}{r_{i}X_{i}^{t_{-1}}}$. $\{r_i\}$ quantifies the relationship strength between the predicted variable (number of task entities) and the independent variables (number of TDMM entities).

%\subsection{Evaluation}
\paragraph{Evaluation.} 
We evaluate the regression model using the $R^2$ measure (coefficient of determination) to assess the goodness of fit. Additionally, we perform a null hypothesis test to determine the statistical significance of the partial regression coefficients.

%\subsection{Results and Discussion}
\paragraph{Results and Discussion.} \noindent
\newline{\bf 1) Optimized Number of Variables.} In our initial experiment, we determine the optimal number of variables and summarize the corresponding $R^2$ values in Table \ref{tab:reg_var}. Additionally, all regression coefficients are statistically significant at $5\%$ level, indicating their strong relationship with the predicted variable. {\bf Discussion:} The overall results indicate that the model achieves a good fit to the data when all four variables (number of tasks, datasets, metrics, and method entities) are used to predict the number of task entities in subsequent years. We also explore the possibility of reducing the number of variables while maintaining similar performance. Interestingly, using only one variable results in a significant drop of 0.1 in the $R^2$ value ($R^2$ value $0.87$), indicating a poor fit to the model. Conversely, increasing the number of variables improves the model fit, suggesting the significance of all four variables in analyzing research trends ($R^2$ value $0.97$). It is worth noting that we exhaustively explored various combinations of variables, including those presented in the table, and consistently obtained similar results. \newline

\noindent{\bf 2) Influence of the Variables.} In the second experiment, we assess the association between the target variable and each independent variable. In Table~\ref{tab:reg_coeff}, we present the regression coefficients corresponding to each entity type. Larger values of regression coefficients indicate a stronger relationship between the target variable and the respective independent variable. {\bf Discussion:} Overall, we note that the gradual emergence of newer tasks has been a driving force behind research progress. However, when we analyze the trends within each year interval, we uncover more nuanced patterns. During the {\it Early Years (1979--1989)}, when NLP was in its nascent stage as an independent research field, the focus was on creating new {\it datasets} to fuel research advancements. In the {\it Formative Years (1990--2002)}, we witnessed the introduction of new {\it methods}, particularly data-driven approaches, which played a crucial role in shaping the field. Subsequently, {\it from 2003 to 2017}, {\it statistical methods} underwent a revolution, and later in the same period, {\it neural network methods} experienced a resurgence, indicating significant shifts in research trends. Now, in the present {\it Deep Learning Era (2018--2022)}, we observe a rapid creation of newer {\it datasets} in a relatively short span of time, driven by the research needs and the data requirements of deep learning models. These highlight %the
key factors influencing research trajectory over time. %the years.

\begin{table}[t]
    \begin{center}
    \resizebox{\columnwidth}{!}{
    \begin{tabular}{l c c c c}
        \toprule
        \multirow{2}{*}{Years} & \multicolumn{4}{c}{Partial Regression Coefficient} \\
        
        \cmidrule(lr){2-5}
        
        & Tasks & Datasets & Methods & Metrics \\
        
        \toprule
        
        1979--1989 &  0.35 & {\bf 2.24} & 0.21 & 0.02\\
        1990--2002 &  0.82 & 0.89 & {\bf 2.86} & 0.81\\
        2003--2017 &  5.37 & 6.26 & {\bf 7.00} & 0.69\\
        2018--2022 &  1.47 & {\bf 3.38} & 1.79 & 0.41\\
        
        \midrule
        
        1979 - 2022 & {\bf 3.50} & 1.07 & 2.92 & 0.54\\
        \bottomrule
        
    \end{tabular}
    }
    % \caption{{\bf Factors Influencing NLP Task Entities:} Overall results show that newer NLP tasks are highly associated with older tasks.}
    \caption{Variables Influencing NLP task entities.}    
    \label{tab:reg_coeff}
    \end{center}
    \vspace*{-3mm}
\end{table}

%\section{Methods Overview}
%\section{Methods}
\section{Causal Methodology for NLP Research Analysis}
\label{sec:method}

Drawing on the insights gained from the Regression Analysis (Section \ref{subsec:mult_reg}), we now establish the cornerstone of our study by defining three causal variables that drive the causal analysis in the subsequent sections. Using causal discovery and inference techniques, we analyze the causal relationships among the variables and measure the impact of TDMM entities on target task entities based on these relationships. Figure \ref{fig:method} illustrates the architecture that underpins our framework.

\subsection{Causal Variables}
\label{subsec:variables}

\paragraph{Task Frequency Shift Value:} Distinguishing from the previous approaches \citep{tan-etal-2017-friendships, prabhakaran-etal-2016-predicting}, that rely on word frequencies, we define task frequency $f(y)_{t}$ as the number of published papers focusing on a specific task $y$ in a given year $t$, normalized by the total number of papers published on the same year. The task frequency shift value $\Delta{freq_{t_1}^{t_2}(y)}$ captures the average change in the number of published papers on $y$ between two years $t_1 < t_2$. This value serves as a measure of the research trends associated with the task during that time interval, indicating whether it experienced growth or decline. The frequency shift value is given by: 
$\Delta{freq_{t_1}^{t_2}(y)} = \frac{f(y)_{t_2} - f(y)_{t_1}}{t_2 - t_1}$.

% \begin{equation}\label{eqn:freq_shift}
%   \Delta{freq_{t_1}^{t_2}(y)} = \frac{f(y)_{t_2} - f(y)_{t_1}}{t_2 - t_1}  
% \end{equation}

\paragraph{Task Stability Value:} We introduce the concept of task stability value to measure the change in the research context of a given task, $y$, between two years, $t_1 < t_2$. This value quantifies the overlap in neighboring TDMM entities that appear in the same publication as $y$ within the specified time interval. To calculate task stability, we adapt the semantic stability approach of \citet{wendlandt-etal-2018-factors} to our setting and define it specifically for task entities. %Initially,
Following \newcite{mondal-etal-2021-end}, 
we represent each paper in our dataset as a sequence of TDMM entity mentions, removing non-entity tokens. We then employ ``Skip-gram with negative sampling'' \citep{MikolovSCCD13} to obtain embeddings from this representation. Formally, let $e_1, e_2, ..., e_n$ be this entity representation of a paper, and the objective of skip-gram is to maximize the mean log probability $\frac{1}{n}\sum_{i=1}^{n}\sum_{-c\leq j \leq c}log p(e_{i+j}|e_{i})$, where $c$ is called the context window size. Finally, the task stability value $\Delta{stability_{t_1}^{t_2}(y)}$ of $y$ between $t_1$ and $t_2$ is computed as the percentage overlap between the nearest $l$ neighboring entities of the given task in two representation spaces. The stability value is given by: $\Delta{stability_{t_1}^{t_2}(y)} = \frac{|{\cal{N}}_{t_1}^{l}(y) \cap {\cal{N}}_{t_2}^{l}(y)|}{|{\cal{N}}_{t_1}^{l}(y) \cup {\cal{N}}_{t_2}^{l}(y)|}$, where ${\cal{N}}_{t}^l(y)$ is the set of $l$ neighbours of $y$ in the representation space of year $t$. In this study, we consider the context window $c$ to encompass the entire document, and we set the value of $l$ to $5$.

\paragraph{Entity Change Value:} 
%To understand a task entity's development,
%We use this variable 
We use entity change value to track emerging and disappearing of specific TDMM entities associated with a task, quantifying these changes and capturing related entity occurrences within a specific time period. 
Put simply, we measure the difference in the co-occurrence frequency of a TDMM entity $x$ and a task $y$ between two years $t_1$ and $t_2$. When we identify a significant change in the co-occurrence frequency of $x$ and $y$ over this period, it likely signals a shift in the relation between $x$ and $y$ and, in turn, a shift in NLP Research trends. 
We define entity change value ${\delta}_{y}(x)_{t_1}^{t_2}$ of an entity $x$ of type $\tau(x) \in$ \{task, dataset, metric, method\} with respect to a task $y$ as the absolute difference in frequencies of $x$ co-occurring with $y$ in the same sentence, between years $t_1$ and $t_2$ normalized by the total number of entities of the same type as $x$ that co-occur with $y$ in both years. The entity change value is given by: ${\delta}_{y}(x)_{t_1}^{t_2} = \frac{|C_{t_1}(x, y) - C_{t_2}(x, y)|}{\sum_{\forall e: \tau(e)=\tau(x)}{(C_{t_1}(e, y) + C_{t_2}(e, y)})}$, where the frequency of $x$ co-occurring with $y$ in year $t$ is given by $C_{t}(x, y)$.

In summary, we quantify task trends and research context changes using \emph{task frequency change} and \emph{task stability values}. Below we explore the relationship between \emph{entity change values} and these two variables and estimate the causal impact of TDMM entities on task research landscapes.

\subsection{Causal Algorithms}

%\subsubsection{Causal Structure Discovery}
\paragraph{Causal Structure Discovery}
To uncover the causal structure among variables from observational data, we employ DirectLinGAM \citep{DirectLingam}, which assumes a non-Gaussian data-generating process. Since the variables in Section~\ref{subsec:variables} come from non-Gaussian frequency distributions, DirectLinGAM is suitable. It uses an entropy-based measure to subtract the effect of each independent variable successively. Unlike PC-Stable \citep{JMLR:v15:colombo14a}, it does not require iterative search or algorithmic parameters. We apply DirectLiNGAM with a $5\%$ significance level for causal discovery (see Appendix \ref{sec:appendix_algo} for details).

\paragraph{Causal Inference}
%\subsubsection{Causal Inference}
Once the causal structure between the variables has been established, we leverage this structure to assess the causal effects. Specifically, we measure the causal effects by the {\it entity change value} of entity $x$ on the {\it frequency shift} and subsequently on the {\it stability values} associated with a given task $y$.  For this purpose, we use the probability density function instead of probability mass, as all our causal variables are continuous in nature. We measure the causal effects in two steps: first, we estimate the probability density of the {\it entity change variable} using a linear regression model. In the next step, we regress the {\it frequency shift} and {\it stability} against the {\it entity change value}, weighted by the inverse probability densities obtained in the previous step. We model the functional form of this regression using a spline to avoid bias due to misspecification. Finally, we calculate the causal effect as \newcite{NEURIPS2020_7d265aa7}: $\mu(\Delta{freq_{t_1}^{t_2}(y)}) = \mathbb{E}[\Delta{freq_{t_1}^{t_2}(y)}| \delta_{y}(x)_{t_1}^{t_2}]$ and similarly, $ \mu(\Delta{stability_{t_1}^{t_2}(y)}) = \mathbb{E}[\Delta{stability_{t_1}^{t_2}(y)}| \delta_{y}(x)_{t_1}^{t_2}]$.

\section{Results and Analysis}
\label{sec:res}

Correlation-based measures provide a simple way to quantify the association between variables. However, they fall short of explaining complex cause-effect relationships and can yield misleading results. Causality is essential for gaining a deeper understanding of variable relationships, enhancing the robustness and reliability of our findings beyond the limitations of correlation. We discuss more about the importance of causal methods over correlation-based measures in Section~\ref{sec:corr_cause}.
%In Appendix~\ref{app:causality_vs_correlation}, we delve into the importance of causal methods over correlation-based measures with more depth. 
In this section, our focus is on uncovering relationships among causal variables (Section~\ref{subsec:causal_discovery}) and measuring the impact of TDMM entities on target task entities (Section~\ref{subsec:causal_inference}).

\subsection{Causal Relation between the Variables}
\label{subsec:causal_discovery}

\begin{figure}
    \centering
    \includegraphics[width=0.5\textwidth]{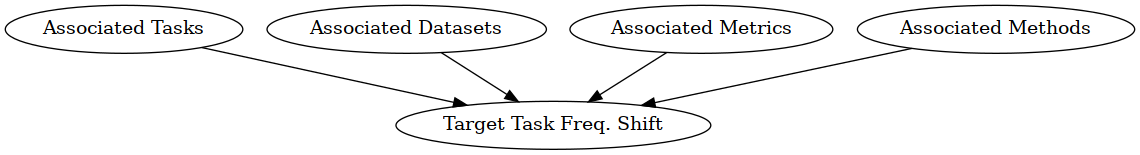}
    \caption{Causal Graph of TDMM Entities (entity change values) and Task Entity Frequency Shift. }
    %\caption{{\bf Causal Graph:} The graph shows that the emergence and disappearance of TDMM entities (entity change values) have a direct causal effect on the frequency shift of task entities.}
    %\vspace*{-3mm}
    \label{fig:causal_relation}
\end{figure}

%To elicit the causal relationship between the task entities and other relevant TDMM  entities in our dataset, we apply DirectLiNGAM \citep{JMLR:v12:shimizu11a} discovery algorithm with $5\%$ significance level (see Appendix \ref{sec:appendix_algo} for more details). 

Figure~\ref{fig:causal_relation} shows the discovered causal graph for the frequency shift of task entities. Overall, we observe that the entity change values of associated tasks, datasets, metrics, and methods have a direct causal effect on the frequency shift values of the target tasks. Since frequency shift value quantifies the trend in NLP research, we infer from the causal graph that the trend of a task is governed primarily by the life cycles of its associated TDMM entities. We see similar causal relation on task stability value (see Figure~\ref{fig:causal_relation_stability}, Appendix~\ref{sec:appendix_inference}). {\bf Evaluation:} We perform a sensitivity analysis of the causal graph by adding Gaussian noise with zero mean and unit variance to the entity change values in the data \citep{pmlr-v97-cinelli19a}. This gives an estimate of the robustness of the graph in the presence of unobserved confounders. We observe that the graph is stable to unobserved confounding, giving all edge probabilities greater than 0.5.

\subsection{Causal Impact of the Variables}
\label{subsec:causal_inference}

\begin{table*}[!ht]
    \centering
    \begin{adjustbox}{width=2.0\columnwidth, center}
    %\resizebox{2.09\columnwidth}{!}{
    \begin{tabular}{l c c c c|r}
    \toprule
    \multirow{4}{*}{\textbf{Task}} & \multicolumn{5}{c}{\textbf{Primary Cause}} \\
    \cmidrule(lr){2-6}
    & \multicolumn{1}{c}{\textbf{1979-1989}} & \multicolumn{1}{c}{\textbf{1990-2002}} & \multicolumn{1}{c}{\textbf{2003-2017}} & \multicolumn{1}{c}{\textbf{2018-2022}} & \multicolumn{1}{c}{\textbf{1979-2022}}\\
    %& Type & Entity & Type & Entity & Type & Entity & Type & Entity & Type & Entity \\
    \toprule
    Language Modeling & - & - & Recurrent Neural Networks$^M$ & Transformers$^M$ & Transformers$^M$ \\ 

   Dialogue System & - & Probabilistic Generative Models$^M$ & Recurrent Neural Networks$^M$ & MultiWoz$^D$ & MultiWoz$^D$ \\

   Machine Translation & - & Probabilistic Generative 
Models$^M$ & WMT Data$^D$ & Transformers$^M$ & Transformers$^M$ \\

    Speech Recognition & Hidden Markov Models$^M$ & Hidden Markov Models$^M$ & Machine Translation$^T$ & Machine Translation$^T$ & Hidden Markov Models$^M$ \\

   \hline

   Named Entity Recognition & - & Hidden Markov Models$^M$ & POS Tagging$^T$ & Relation Extraction$^T$ & POS Tagging$^T$ \\

    POS Tagging & - & Text Classification$^T$ & Parser Algorithms$^M$ & Word Segmentation$^T$ & Word Segmentation$^T$ \\

   Semantic Parsing & Grammar Induction$^M$ & Parser Algorithms$^M$ & Parser Algorithms$^M$ & Dependency Parsing$^T$ & Parser Algorithms$^M$ \\
    
    %Word Sense Disambiguation & - & Wordnet$^D$ & Maximum Entropy Models$^M$ & Neural Network Models$^M$ & Wordnet$^D$ \\

  Morphological Analysis & - & Statistical Models$^M$ & Dependency Parsing$^T$ & UD Treebank$^D$ & Statistical Models$^M$ \\

    \hline

    %Word Sense Disambiguation & - & Wordnet$^D$ & Maximum Entropy Models$^M$ & Neural Network Models$^M$ & Wordnet$^D$\\
    
    %\hline

    %Semantic Parsing & Grammar Induction$^M$ & Parser Algorithms$^M$ & Parser Algorithms$^M$ & Dependency Parsing$^T$ & Parser Algorithms$^M$ \\
    
    Semantic Role Labeling & - & - & Support Vector Machines$^M$ & Neural Network Models$^M$ & Support Vector Machines$^M$ \\

        Co-reference Resolution & - & MUC-VI Text Collection$^D$ & Integer Linear Programming$^M$ & Neural Network Models$^M$ & Neural Network Models$^M$ \\

\hline

    Word Sense Disambiguation & - & Wordnet$^D$ & Maximum Entropy Models$^M$ & Neural Network Models$^M$ & Wordnet$^D$\\
    
    \hline

 Sentiment Analysis & - & - & Twitter Dataset$^D$ & Text Classification$^T$ & Text Classification$^T$ \\
    
    Argument Mining & - & - & Text Classification$^T$ & Sentiment Analysis$^T$ & Sentiment Analysis$^T$ \\

   \hline
   
    Question Answering & Parsing Algorithms$^M$ & Information Extraction$^T$ & Information Extraction$^T$ & Pre-Trained LLMs$^M$ & Information Extraction$^T$ \\

   Textual Entailment & - & - & Stastical Models$^M$ & Pre-Trained LLMs$^M$ & Pre-Trained LLMs$^M$ \\

    Summarization & - & Wordnet$^D$ & Sentence Compression$^T$ & Pre-Trained LLMs$^M$ & Pre-Trained LLMs$^M$ \\

    \bottomrule
    \end{tabular}
    \end{adjustbox}
    %}
    \caption{Causal analysis identifies the main drivers (Methods, Tasks, Datasets) of frequency shifts in NLP tasks across four periods, with "-" indicating insufficient data for analysis.}
    %\caption{{\bf The primary reason behind the frequency shift of the tasks.} We analyze the trends in four different periods of reference. Overall, we observe that Methods (M), Tasks (T), and Datasets (D) are the primary causes behind the paradigm shift of these 16 NLP tasks from different research areas. %NLP research areas. 
    %"-" means there is not enough data instances for the causal analysis. %For details, refer to Section~\ref{subsec:causal_inference}.
    %}
    \vspace*{-3mm}
    \label{tab:freq_change_cause}
\end{table*}

The organizers of ACL 2018\footnote{\url{https://acl2018.org/call-for-papers/}} categorize NLP research into 21 areas, and %describes each area as a set of tasks, methods, datasets, and metrics 
provide a set of popular tasks for each area. Out of those, we curate 16 areas and select one task from each based on its frequency of occurrence in our corpus. We estimate the effect of TDMM entities (entity change value) behind the development of these tasks (frequency shift value) (see Section~\ref{subsec:variables}) and summarize the results in Table~\ref{tab:freq_change_cause}. Since we do not have confounders (Section~\ref{subsec:causal_discovery}), evaluating the causal effect reduces to estimating the conditional expectation of the frequency shift values given the entity change values. We present detailed results in Table~\ref{tab:freq_change_cause_additional}. We examine the results by addressing the following set of inquiries.\\

% \subsubsection{What role does the emergence and adoption of new methodologies play in causally driving the shift in NLP tasks?}
\noindent
\begin{enumerate*}[label={Q\arabic*.}]
%\item {\bf What role does the emergence and adoption of new methodologies play in causally driving the shift in NLP tasks?}
\item {\bf What role do the methodologies play in causally driving the shift in NLP tasks?}
\end{enumerate*}

\noindent New methodologies have a significant influence on research in various areas of Natural Language Processing (NLP). In the field of Language Modeling, we observe a shift in influence between different methodologies over time. 

Between 2003 and 2017, Recurrent Neural Networks (RNNs) had the most decisive impact on {\bf Language Modeling} research. However, this trend shifted with the emergence of Transformers, which have since become the dominant influence in research on this task.

{\bf Dialogue Systems}, which involve automatic response generation, are closely related to Language Modeling. Therefore, research in this area is highly influenced by Generative Models. From 1990 to 2002, Probabilistic Models played a crucial role in shaping Dialogue Systems research, while RNNs took the lead between 2003 and 2017.

{\bf Machine Translation}, another task related to Language Modeling, requires the generation of the translated text. Naturally, we observe the influence of similar entities in Machine Translation research. Probabilistic Models had the most decisive impact between 1990 and 2002. In recent years (2018-2022), Transformers have emerged as the dominant influence in this research area.

In the field of {\bf Speech Recognition}, Hidden Markov Models (HMMs) have shown a significant influence. HMMs have played a crucial role in shaping Speech Recognition research between 1979 to 2002.

{\bf Named Entity Recognition} (NER) has also been influenced by Hidden Markov Models, particularly in its early days (1990-2002), as NER is often formulated as a sequence tagging problem. Various parser algorithms were employed to solve the problem in the period between 2003 and 2017.

For {\bf Semantic Parsing}, parser algorithms have been instrumental and have had a significant impact on research in this area. Between 1979 and 1989, Grammar Induction techniques were used to elicit the underlying semantic parse trees.

From 1990 to 2002, researchers employed various statistical models in {\bf Morphological Analysis}, which is evident from our results.

In {\bf Semantic Role Labeling}, Support Vector Machines and Neural Network Models have been widely used to solve this task.

In {\bf Co-reference Resolution}, Neural Network models have gained prominence starting in 2018. However, from 2003 to 2017, Integer Linear Programming was also utilized to address this problem.

Pre-trained Language Models (LLMs) have demonstrated superior performance in several NLP tasks, including {\bf Question Answering}. Researchers have also explored parsing algorithms to parse questions and align them with potential answers.

Furthermore, {\bf Textual Entailment} and {\bf Summarization} have been heavily influenced by pre-trained LLMs between 2018 and 2022, as evident from our results. \\

\noindent
\begin{enumerate*}[label={Q\arabic*.}, resume]
\item {\bf How have changes in data availability contributed to the NLP Research Tasks?}
\end{enumerate*}
%\subsubsection{How have changes in data availability contributed to the NLP Research Tasks?}

\noindent High-quality datasets play a crucial role in advancing NLP research. While new methodologies are important, they cannot fully propel the field forward without the support of high-quality datasets. Researchers understand the significance of dataset quality and actively curate datasets to drive advancements in the field. Our findings further confirm the prevalence of this trend, highlighting the strong emphasis on dataset quality in NLP research.

In the early stages of deep neural models, such as Recurrent Neural Networks (RNNs), the creation of large datasets became essential for efficient model training. Between 2018 and 2022, several datasets were curated, with MultiWoz being the most widely used dataset for research in {\bf Dialogue Systems}.

In the domain of {\bf Machine Translation}, the significance of datasets in shaping research direction cannot be overlooked. The influence of WMT datasets on Machine Translation research is evident from our findings.

For {\bf Morphological Analysis}, the Universal Dependency Treebank dataset is frequently used as a benchmark, indicating its importance in driving research in this area.

During the period of 1990-2002, the creation of the MUC-VI dataset played a crucial role in advancing research in {\bf Co-reference resolution}.

In the field of {\bf Sentiment Analysis}, the Twitter dataset holds significant importance in driving research in this domain.

Overall, our analysis underscores the vital role of datasets in shaping and driving research across various NLP tasks. \\

\noindent
\begin{enumerate*}[label={Q\arabic*.}, resume]
%\item {\bf Are there any causal relationships between the adoption of specific metrics for evaluating NLP performance and the subsequent paradigm shift in research directions?}
\item {\bf Do evaluation metrics drive paradigm shifts in NLP research?}
\end{enumerate*}
%\subsubsection{Are there any causal relationships between the adoption of specific metrics for evaluating NLP performance and the subsequent paradigm shift in research directions?}

\noindent Most NLP tasks rely on a standard set of metrics borrowed from other domains, such as machine learning and computer vision, to evaluate system performance. However, there is limited research dedicated to improving these metrics within the field of NLP, as it often requires theoretical knowledge beyond the scope of NLP itself. Despite this, our analysis in Table~\ref{tab:freq_change_cause_additional} reveals some noteworthy exceptions. Metrics explicitly designed for evaluating NLP tasks, such as BLEU and METEOR, have demonstrated significant impact in advancing {\bf Machine Translation} research. Similarly, the metric ROUGE has influenced research in the field of {\bf Summarization}. While perplexity scores are commonly used to measure the generalization capabilities of probability distributions, they are predominantly utilized for evaluating language models in NLP tasks. \\

\noindent
\begin{enumerate*}[label={Q\arabic*.}, resume]
\item {\bf What is the causal impact of cross-pollination of ideas between related NLP tasks?}
\end{enumerate*}
%\subsubsection{What is the causal impact of cross-pollination of ideas between related NLP tasks?}

\noindent We consistently observe a pattern of related NLP tasks evolving in tandem, borrowing ideas and techniques from one another. This trend is clearly reflected in our findings. For instance, {\bf Speech Recognition} and {\bf Machine Translation} are linked as researchers explore end-to-end systems that translate speech, and our results show that Machine Translation has had the greatest influence on Speech Recognition research between 2003 and 2022.

{\bf Named Entity Recognition} (NER) is commonly approached as a sequence tagging problem, and it is influenced by related tasks such as {\bf POS Tagging} (2003-2017) and {\bf Relation Extraction} (2018-2022), as these problems are often jointly solved. Similarly, POS Tagging initially posed as a text classification problem (1990-2002), is significantly impacted by the {\bf word segmentation task}, as evident from our results in the period of 2018-2022.

In recent years (2018-2022), {\bf dependency} and {\bf semantic parsing} have been jointly solved using the same neural model, highlighting the influence of dependency parsing on research in semantic parsing. {\bf Sentiment Analysis} has garnered considerable research interest and is commonly framed as a text classification problem. Additionally, {\bf Argument Mining}, which involves understanding the sentiments behind arguments, is influenced by sentiment analysis. Furthermore, the classification of various argument components, such as claims and evidence, is often approached as text classification problems, as evidenced by our results.

\section{Discussion: Correlation and Causation}
\label{sec:corr_cause}

\epigraph{``correlation does not imply causation''}{-- \citet{pearson1892grammar}}

Causation and correlation, although related, are distinct concepts. While they can coexist, correlation does not simply imply causation. Causation signifies a direct cause-and-effect relationship, where one action leads to a specific outcome. In contrast, correlation simply indicates that two actions are related in some way, without one necessarily causing the other. 

In our work, we focus on causal inference from data. While correlation-based measures provide a straightforward method for quantifying associations between variables, they often fall short when it comes to explaining complex cause-and-effect relationships. 

To demonstrate the effectiveness of our framework, we establish a simple baseline using a PMI-based correlation measure ~\citep{bouma2009normalized}. For this analysis, we select {\bf Machine Translation} as our target task entity due to its prominent presence in our corpus and the NLP research landscape. We calculate the PMI scores of {\bf Machine Translation} with all other TDMM entities. The PMI score represents the probabilities of co-occurrence between two entities in sentences from research papers, normalized by their individual occurrence probabilities.

Interestingly, we find that {\bf accuracy}, an entity of type {\bf metric}, has the highest PMI score with Machine Translation among all other entities. However, it is important to note that accuracy is a widely used metric across various NLP tasks, and it is not specifically developed for machine translation, nor has machine translation influenced the concept of accuracy. This observation emphasizes the insufficiency of relying solely on correlation-based metrics to understand and analyze research influence on an entity.

We observe that relying solely on correlations can lead to misleading results and interpretations. Therefore, in order to understand the influence of associated TDMM entities on NLP Task entities, we utilize causal algorithms that enable us to gain insights into the cause-and-effect dynamics among the variables we study.

%\section{Conclusion}
\section{Concluding Remarks}
In this paper, we retrospectively study NLP research from a causal perspective, quantifying research trends of task entities and proposing a systematic framework using causal algorithms to identify key reasons behind the emergence or disappearance of NLP tasks. Our analysis reveals that tasks and methods are the primary drivers of research in NLP, with datasets following their influence, while metrics have minimal impact. It is important to note that in our analysis, we have structured the reference time into four distinct intervals (see Table~\ref{tab:year_intervals}); however, it can be applied to diverse timeframes, ranging from longer periods to brief intervals, including single years. This adaptability, in the context of rapid recent advancements in NLP, allows to zoom in on local trends and developments that might otherwise go unnoticed (such as the influence of in-context learning on NLP tasks). 

We believe our causal analysis enhances understanding of the interplay of research entities in NLP, contributing to the growing body of work on causality and NLP \citep{DBLP:journals/corr/abs-2109-00725}. We provide with additional analysis and insights in Appendix~\ref{app:supp_analysis}.

\section*{Limitations}

This work is centered on NLP research papers from ACL Anthology, with a focus on papers from the ``ACL Events'' category. The ``ACL Events'' category encompasses major conferences, workshops, and journals, including ACL, NAACL, EMNLP, EACL, AACL, CL, and TACL. We also include papers published at COLING from the ``non-ACL Events'' category. Nevertheless, it is important to acknowledge the presence of NLP papers beyond ACL Anthology in AI journals, regional conferences, and preprint servers. Furthermore, we recognize that certain NLP papers may become available on preprint servers before their official publication in peer-reviewed venues. In this study, we focus on ACL Anthology, which can introduce a time lag when assessing the early impact of influential papers released as preprints (e.g., BERT) or only on preprint servers (e.g., RoBERTa). To address such challenges, we leave the curation and inclusion of NLP research papers from these alternative sources for future works.

Our framework requires research papers tagged with entities as input. Hence, the quality of the tags plays a crucial role in the causal inference of our proposed method. The taggers generate noisy outputs and, thus, might require human intervention to denoise the tags. Moreover, causal algorithms require a large amount of data to produce statistically significant results. Hence, research areas that are less explored or newly emerging may not always be suitable for this framework to be applied. Additionally, we highlight that in this work, we do not consider extra-linguistic factors like author affiliations, funding, gender, etc. We leave them for future research work.

\section*{Ethics Statement}

In this work, we use publicly available data from ACL Anthology and do not involve any personal data. It is important to recognize that, while our framework is data-driven, individual perspectives toward research are inherently subjective. Decisions involving science should consider data as well as ethical, social, and other qualitative factors. Furthermore, we underscore that the low influence of TDMM entities in our analysis should not be the sole reason for devaluing research papers or reducing their investments. Ethical and academic considerations should guide decisions on research evaluation and resource allocation.

\section*{Acknowledgements}

We thank Ilia Kuznetsov for his feedback on the initial version of this work. We appreciate all the anonymous reviewers for their helpful comments and suggestions for further analysis. This work has been funded by the German Research Foundation (DFG) as part of the Research Training Group KRITIS No. GRK 2222.

\bibliography{anthology,custom}
\bibliographystyle{acl_natbib}

\appendix

%\section{Appendix: Data}
%\label{sec:appendixa}

\section{Appendix: Supplementary Results}
\label{sec:appendix_inference}

\subsection{Causal Relation}

\begin{figure*}
    \centering
    \includegraphics[width=1\textwidth]{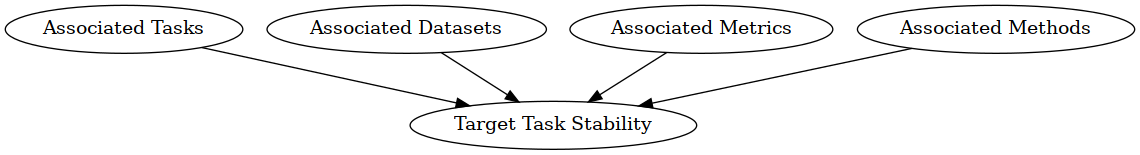}
    \caption{{\bf Causal Graph:} The graph shows that the emergence and disappearance of TDMM entities (entity change values) have a direct causal effect on the stability of task entities. 
    % Refer to Section~\ref{subsec:causal_discovery} for details.
    }
    \label{fig:causal_relation_stability}
\end{figure*}

In Figure~\ref{fig:causal_relation_stability}, we observe that entity change values of tasks, datasets, metrics and methods have direct causal influence on task stability value.

\begin{table*}[!ht]
    \centering
    \resizebox{2.08\columnwidth}{!}{
    \begin{tabular}{l c c c c|r}
    \toprule
    \multirow{4}{*}{\textbf{Task}} & \multicolumn{5}{c}{\textbf{Primary Cause}} \\
    \cmidrule(lr){2-6}
    & \multicolumn{1}{c}{\textbf{1979-1989}} & \multicolumn{1}{c}{\textbf{1990-2002}} & \multicolumn{1}{c}{\textbf{2003-2017}} & \multicolumn{1}{c}{\textbf{2018-2022}} & \multicolumn{1}{c}{\textbf{1979-2022}}\\
    %& Type & Entity & Type & Entity & Type & Entity & Type & Entity & Type & Entity \\
    \toprule
    \multirow{4}{*}{Language Modeling}  & - & - & Recurrent Neural Networks$^M$ & Transformers$^M$ & Transformers$^M$ \\
    
     & - & - & Machine Translation$^T$ & Text Generation$^T$ & Text Generation$^T$ \\
     
      & - & - & Penn Treebank$^D$ & Perplexity$^m$ & Perplexity$^m$ \\
      
      & - & - & Perplexity$^m$ & SuperGLUE$^D$ & Penn Treebank$^D$ \\
      
      \midrule
      
      \multirow{4}{*}{Dialogue System} & - & - & Recurrent Neural Networks$^M$ & MultiWoz$^D$ & MultiWoz$^D$ \\
      
      & - & - & MultiWoz$^D$ & Transformers$^M$ & Transformers$^M$ \\
      
      & - & - & Language Generation$^T$ & Response Generation$^T$ & Response Generation$^T$ \\
      
      & - & - & Perplexity$^m$ & Rouge$^m$ & Rouge$^m$ \\
      
      \midrule
      
    \multirow{4}{*}{Machine Translation} & - & Probabilistic Generative 
    Models$^M$ & WMT Data$^D$ & Transformers$^M$ & Transformers$^M$ \\
    
    & - & Speech Recognition$^T$ & BLEU$^m$ & METEOR$^m$ & METEOR$^m$ \\
    
    & - & Perplexity$^m$ & Attention Mechanism$^M$ & Language Modeling$^T$ & Language Generation$^T$ \\
    
    & - & Penn Treebank$^D$ & Language Generation$^T$ & WMT Data$^D$ & WMT Data$^D$ \\
    
    \midrule
    
    \multirow{4}{*}{Speech Recognition} & Hidden Markov Models$^M$ & Hidden Markov Models$^M$ & Machine Translation$^T$ & Machine Translation$^T$ & Hidden Markov Models$^M$ \\
    
    & Machine Translation$^T$ & WSJ Corpus $^D$ &Hidden Markov Models$^M$ & Acoustic Models $^M$ & Language Modeling$^T$ \\
    
    & Perplexity$^m$ & Perplexity$^m$ & ATIS Dataset$^D$ & Switchboard Dataset$^D$ & Perplexity$^m$ \\
    
    & -  & Language Modeling$^T$ & Word Error Rate$^m$ & Word Error Rate$^m$ & ATIS Dataset$^D$ \\
      
    \midrule
    \midrule
    
    \multirow{4}{*}{Named Entity Recognition} & - & Hidden Markov Models$^M$ & POS Tagging$^T$ & Relation Extraction$^T$ & POS Tagging$^T$ \\
    
    & - & Information Extraction$^T$ & Conditional Random Fields$^M$ & Wikipedia Corpus$^D$ & Random Fields$^M$ \\
    
    & - & Genia Corpus$^D$ & Pubmed$^D$ & Pre-Trained LLMs$^M$ & Conditional Ontonotes$^D$ \\
    
    & - & F1 Score$^m$ & F1 Score$^m$ & F1 Score$^m$ & F1 Score$^m$ \\
    
    \midrule

    \multirow{4}{*}{POS Tagging} & - & Text Classification$^T$ & Parser Algorithms$^M$ & Word Segmentation$^T$ & Word Segmentation$^T$ \\
    
    & - & Discriminative Models$^M$ & Word Segmentation$^T$ & Neural Network Models$^M$ & Neural Network Models$^M$ \\
    
    & - & Penn Treebank$^D$ & Penn Treebank$^D$ & Penn Treebank$^D$ & Penn Treebank$^D$ \\
    
    & - & F1 Score$^m$ & F1 Score$^m$ & F1 Score$^m$ & F1 Score$^m$ \\
    
    \midrule

    \multirow{4}{*}{Word Sense Disambiguation} & - & Wordnet$^D$ & Maximum Entropy Models$^M$ & Neural Network Models$^M$ & Wordnet$^D$ \\
    
    & - & Semantic Tagging$^T$ & Text Classification$^T$ & Text Classification$^T$ & Neural Network Models$^M$ \\
    
    & - & Discriminative Models$^M$ & Wordnet$^D$ & Wordnet$^D$ & Text Classification$^T$ \\
    
    & - & Accuracy$^m$ & F1 Score$^m$ & F1 Score$^m$ & F1 Score$^m$ \\
    
    \midrule

    \multirow{4}{*}{Morphological Analysis} & - & Statistical Models$^M$ & Dependency Parsing$^T$ & UD Treebank$^D$ & Statistical Models$^M$ \\
    
    & - & Word Segmentation$^T$ & Statistical Models$^M$ & Pre-Trained LLMs$^M$ & Dependency Parsing$^T$ \\
    
    & - & UD Treebank$^D$ & UD Treebank$^D$ & Lemmatization$^T$ & UD Treebank$^D$ \\
    
    & - & Accuracy$^m$ & Accuracy$^m$ & F1 Score$^m$ & Accuracy$^m$ \\
    
    \midrule
    \midrule
    
    \multirow{4}{*}{Semantic Parsing} & Grammar Induction$^M$ & Parser Algorithms$^M$ & Parser Algorithms$^M$ & Dependency Parsing$^T$ & Parser Algorithms$^M$ \\
    
    & Information Retrieval$^T$ & Information Extraction$^T$ & Dependency Parsing$^T$ & Parser Algorithms$^M$ & Penn Treebank$^D$ \\
    
    & Accuracy$^m$ & Penn Treebank$^D$ & Penn Treebank$^D$ & Penn Treebank$^D$ & Dependency Parsing$^T$ \\
    
    & - & F1 Score$^m$ & F1 Score$^m$ & F1 Score$^m$ & F1 Score$^m$ \\
    
    \midrule
    
    \multirow{4}{*}{Semantic Role Labeling} & - & - & Support Vector Machines$^M$ & Neural Network Models$^M$ & Support Vector Machines$^M$ \\
    
    & - & - & Relation Extraction$^T$ & Named Entity Recognition$^T$ & Named Entity Recognition$^T$ \\
    
    & - & - & Propbank$^D$ & Propbank$^D$ & Propbank$^D$ \\
    
    & - & - & F1 Score$^m$ & F1 Score$^m$ & F1 Score$^m$ \\
    
    \midrule

        \multirow{4}{*}{Co-reference Resolution} & - & MUC-VI Text Collection$^D$ & Integer Linear Programming$^M$ & Neural Network Models$^M$ & Neural Network Models$^M$ \\
        
        & - & Discriminator Models$^M$ & Ontonotes$^D$ & Ontonotes$^D$ & Ontonotes$^D$ \\
        
        & - & Word Sense Disambiguation$^T$ & Mention Detection$^T$ & Mention Detection$^T$ & F1 Score$^m$ \\
        
        & - & F1 Score$^m$ & F1 Score$^m$ & F1 Score$^m$ & Mention Detection$^T$ \\
        
    \midrule
    \midrule
    
     \multirow{4}{*}{Sentiment Analysis} & - & - & Twitter Dataset$^D$ & Text Classification$^T$ & Text Classification$^T$ \\
     
     & - & - & Text Classification$^T$ & Pre-Trained LLMs$^M$ & Neural Network Models$^M$ \\
     
     & - & - & Neural Netowrk Models$^M$ & Amazon Reviews$^D$ & Twitter Dataset$^D$ \\
     
     & - & - & F1 Score$^m$ & F1 Score$^m$ & F1 Score$^m$ \\
     
     \midrule
    
    \multirow{4}{*}{Argument Mining} & - & - & Text Classification$^T$ & Sentiment Analysis$^T$ & Sentiment Analysis$^T$ \\
    
    & - & - & Neural Network Models$^M$ & Neural Network Models$^M$ & Neural Network Models$^M$ \\
    
    & - & - & Wikipedia Corpus$^D$ & Wikipedia Corpus$^D$ & Wikipedia Corpus$^D$ \\
    
    & - & - & F1 Score$^m$ & F1 Score$^m$ & F1 Score$^m$ \\
    
    \midrule
    \midrule
    
    \multirow{4}{*}{Question Answering} & Parsing Algorithms$^M$ & Information Extraction$^T$ & Information Extraction$^T$ & Pre-Trained LLMs$^M$ & Information Extraction$^T$ \\
    
    & Information Retrieval$^T$ & Wordnet$^D$ & Freebase$^D$ & Squad$^D$ & Pre-Trained LLMs$^M$ \\
    
    & Accuracy$^m$ & Accuracy$^m$ & Parsing Algorithms$^M$ & Summarization$^T$ & Squad$^D$ \\
    
    & - & Statistical Models$^M$ & F1 Score$^m$ & F1 Score$^m$ & F1 Score$^m$ \\
    
    \midrule

   \multirow{4}{*}{Textual Entailment} & - & - & Statistical Models$^M$ & Pre-Trained LLMs$^M$ & Pre-Trained LLMs$^M$ \\
   
   & - & - & Information Extraction$^T$ & SNLI Dataset$^D$ & SNLI Dataset$^D$ \\
   
   & - & - & F1 Score$^m$ & Text Classification$^T$ & Text Classification$^T$ \\
   
   & - & - & - & F1 Score$^m$ & F1 Score$^m$ \\
   
   \midrule 

    \multirow{4}{*}{Summarization} & - & Wordnet$^D$ & Sentence Compression$^T$ & Pre-Trained LLMs$^M$ & Pre-Trained LLMs$^M$ \\
    
    & - & Probabilistic Generative Models$^M$ & Recurrent Neural Networks$^M$ & Rouge$^m$ & Rouge$^m$ \\
    
    & - & F1 Score$^m$ & Rouge$^m$ & Pubmed$^D$ & Question Answering$^T$ \\
    
    & - & Information Retrieval$^T$ & Gigaword$^D$ & Question Answering$^T$ & Pubmed$^D$ \\
    
    \bottomrule
    \end{tabular}
    }
    \caption{{\bf The primary reason behind the frequency shift of the tasks.} We analyze the trends in four different periods of reference. Most influential Task(T), Dataset(D), Method(M) and Metric(m) are given in the decreasing order of their influence.   %NLP research areas. 
    "-" means there is not enough data instances for the causal analysis.}
    \label{tab:freq_change_cause_additional}
\end{table*}

\section{Appendix: Supplementary Analysis}
\label{app:supp_analysis}

In addition to the primary results presented in the paper (Section~\ref{sec:res}), in this section, we describe the supplementary analysis. %\\

\subsection{NLP Tasks and Their Dataset Evolution}
\label{app:add_analysis}

\paragraph{Frequently Pursued NLP Tasks.} From Table~\ref{tab:freq_change_cause_additional} in our paper, we observe that overall (from 1979-2022), among all the tasks, “Text Classification” (column 6) holds a remarkable position. This prominence stems from the frequent usage of various NLP tasks being framed or aligned as “Text Classification” or borrowing concepts from it to address other tasks such as “Sentiment Analysis” or “Word Sense Disambiguation.” Additionally, our framework offers the flexibility to perform a similar analysis between any chosen periods.

\paragraph{Evolution of Datasets in NLP Tasks.} Referring to Table~\ref{tab:freq_change_cause_additional} in our paper, in the context of “Speech Recognition,” we observe a shift in influential datasets over different periods. Between 1990-2002, the ``WSJ Corpus'' took the lead, while in the subsequent period of 2003-2017, the ``ATIS Dataset'' had more influence. Interestingly, between 2018-2022, the trend shifted once again to the ``Switchboard Dataset''. 

A similar trend is reflected in the “Summarization” task as well: in the years 1990-2002, “Wordnet” played a significant role, while the “Gigaword Dataset” took over in 2003-2017. However, in the most recent period of 2018-2022, “Pubmed” emerged as the notable dataset for the “Summarization” task.

\paragraph{Common Datasets Across NLP Tasks.} We observe from Table~\ref{tab:freq_change_cause_additional} (column 6) that across the entire span from 1979 to 2022, the “Penn Treebank” dataset emerged as a pivotal influence, significantly impacting tasks such as “Language Modeling,” “POS Tagging,” and “Semantic Parsing.” Using our framework, a similar analysis could also be done between any chosen periods.

\subsection{Entitiy Influence on Task Frequency and Stability}
\label{app:ent_inf}

\paragraph{Influence of Research Entities on Task Stability.} We measure the causal effect of research entities on Task Stability Value (see Section~\ref{subsec:variables}). From the resulting causal graph (Figure~\ref{fig:causal_relation_stability}), we observe that the entity change values of associated tasks, datasets, metrics, and methods directly impact the stability value of the target task, similar to the task frequency shift value.

\paragraph{Correlations Between Task Frequency Change and Stability.} We observe a slightly positive correlation between frequency change and stability of research tasks with a Pearson coefficient of 0.08. This is because when a new task emerges, initially, a few researchers start working on it, which gradually increases its frequency of appearance. At the same time, researchers experiment with various methods and datasets to solve these newly emerged tasks, causing high instability (e.g., Math Problem Solving~\citep{DBLP:journals/corr/abs-1808-07290}). On the contrary, the opposite is not always true: well-defined tasks are often the most researched, and yet researchers always explore new ideas on these tasks, which harms stability.

\section{Appendix: Multiple Linear Regression}
\label{sec:mult_lin_reg}

We use multiple linear regression to regress a variable on several variables~\citep{pearl2016causal}. For instance, if we want to predict the value of a variable $Y$ using the values of variables $X_1, X_2, ..., X_{k-1}, X_k$, we perform multiple linear regression of $Y$ on $\{X_1, X_2, ..., X_{k-1}, X_k\}$, and estimate a regression relationship (Eqn.~\ref{eqn:lin_reg}), which represents an inclined plane through the $(k+1)$-dimensional coordinate system. 

\begin{equation}
\label{eqn:lin_reg}
    Y = r_{0} + \sum_{i=1}^{k}r_{i}X_{i}
\end{equation}

The Gauss-Markov theorem \citep{williams2006gaussian} simplifies the computation of partial regression coefficients ($r_{1}, ..., r_{k}$ in Eqn~\ref{eqn:lin_reg}). It states that if we write Y as a linear combination of $X_{1}, X_{2}, ..., X_{k-1}, X_k$ and noise term $\epsilon$, 

\begin{equation}
    Y = r_{0} + \sum_{i=1}^{k}r_{i}X_{i} + \epsilon
\end{equation}

then, regardless of the distributions of the variables $Y, X_1, X_2, ..., X_k$, the best least-square coefficients are obtained when $\epsilon$ is uncorrelated with each regressors, i.e.,

\begin{equation}
    Cov(\epsilon, X_i) = 0,  \forall i=1, 2, ..., k
\end{equation}

\section{Appendix: Algorithms}
\label{sec:appendix_algo}
\subsection{DirectLinGAM}

%to declare the command name to use first:

%% This declares a command \Comment
%% The argument will be surrounded by /* ... */
\SetKwComment{Comment}{/* }{ */}

\begin{algorithm}
\caption{Causal Graph Discovery: DirectLinGAM-Algorithm}\label{alg:DirectLinGAM}

Given a p-dimensional random vector $x$, a set of its variable subscripts $U$ and a $p \times n$ data matrix of the random vector as $X$, initialize an ordered list of variables $K := \phi$ and $m := 1$\;

Repeat until $p$ - $1$ subscripts are appended to $K$: Perform least square regression of $x_i$ and $x_j$, $\forall i \in U-K(i \neq j)$ and compute the residual vectors $r^{(j)}$ and the residual data matrix $R^{(j)}$ from the matrix $X$, $\forall j \in U-K$. Find a variable $x_m$ independent of its residuals and append $m$ to the end of $K$\;

Append the remaining variable to the end of $K$\;

Construct a strictly lower triangular matrix $B$ by following the order in $K$, and estimate the
connection strengths $b_{ij}$ by using some conventional covariance-based regression such as
least squares and maximum likelihood approaches on the original random vector $x$ and the
original data matrix $X$\;

\end{algorithm}

In Algorithm \ref{alg:DirectLinGAM}, we describe the DirectLinGAM algorithm (oracle version) in high level as described by ~\citet{DirectLingam}.

\end{document}